\documentclass[conference]{IEEEtran}
\IEEEoverridecommandlockouts
\usepackage{cite}
\usepackage{amsmath,amssymb,amsfonts}
\usepackage{algorithmic}
\usepackage{graphicx}
\usepackage{booktabs}
\usepackage{float}
\usepackage{textcomp}
\usepackage{xcolor}
\usepackage{hyperref}
\usepackage{fancyhdr} 
\usepackage{lipsum} 
\hypersetup{
        colorlinks=true, linkcolor=black
    }
\def\BibTeX{{\rm B\kern-.05em{\sc i\kern-.025em b}\kern-.08em
    T\kern-.1667em\lower.7ex\hbox{E}\kern-.125emX}}

\makeatletter
\newcommand{\linebreakand}{%
  \end{@IEEEauthorhalign}
  \hfill\mbox{}\par
  \mbox{}\hfill\begin{@IEEEauthorhalign}
}
\makeatother

\begin{document}

\title{HandGCAT: Occlusion-Robust 3D Hand Mesh Reconstruction from Monocular Images
\thanks{$^{*}$ Corresponding author}
}

\author{Shuaibing Wang$^{1,2}$ $\quad$
        Shunli Wang$^{1,2}$ $\quad$
        Dingkang Yang$^{1,2}\quad$ \\
        Mingcheng Li$^{1,2}\quad$
        Ziyun Qian$^{1,2}\quad$ 
        Liuzhen Su$^{1,2}\quad$ 
        Lihua Zhang$^{1,2,3,4*}$ \\ 
        $^1$Academy for Engineering and Technology, Fudan University$\quad$
        $^2$Institute of Meta-Medical, IPASS\\
        $^3$Jilin Provincial Key Laboratory of
Intelligence Science and Engineering, Changchun, China\\
$^4$AI \& Unmanned Systems Engineering Research Center of Jilin Province, China\\
{\tt\small 21210860016@m.fudan.edu.cn, lihuazhang@fudan.edu.cn}
}



\maketitle

\begin{abstract}
We propose a robust and accurate method for reconstructing 3D hand mesh from monocular images.
This is a very challenging problem, as hands are often severely occluded by objects.
Previous works often have disregarded 2D hand pose information, which contains hand prior knowledge that is strongly correlated with occluded regions.
Thus, in this work, we propose a novel 3D hand mesh reconstruction network HandGCAT, that can fully exploit hand prior as compensation information to enhance occluded region features. Specifically, we designed the Knowledge-Guided Graph Convolution (KGC) module and the Cross-Attention Transformer (CAT) module.
KGC extracts hand prior information from 2D hand pose by graph convolution.
CAT fuses hand prior into occluded regions by considering their high correlation.
Extensive experiments on popular datasets with challenging hand-object occlusions, such as HO3D v2, HO3D v3, and DexYCB demonstrate that our HandGCAT reaches state-of-the-art performance.
The code is available at \href{https://github.com/heartStrive/HandGCAT}{https://github.com/heartStrive/HandGCAT}.
\end{abstract}


\section{Introduction}
3D hand mesh reconstruction plays an important role in many applications, such as virtual reality, human-computer interaction, sign language translation, and robotics \cite{inter_two_hand, i2uv_handnet}. It is now an increasingly popular task in computer vision, benefiting from advancements in deep learning algorithms \cite{yang2022disentangled, yang2022learning,yang2022emotion,yang2023target,kuang2023towards,wang2023model,yang2022contextual,yang2023context,wang2021tsa,wang2022spacenet,wang2021survey}.
As we all know, the RGB camera is lower cost and computationally friendly, and it is highly desired to recover 3D hand mesh from a single RGB image \cite{inter_two_hand,keypoint_transformer}.
Self-occlusion in hand is an inherent problem caused by the diversity of hand poses and shapes. When interacting with objects, hands exhibit even greater occlusion from almost any point of view \cite{collaborative}.
However, eliminating severe occlusion is challenging and remains unsolved.
\begin{figure}
  \centering
  \includegraphics[width=0.8\linewidth]{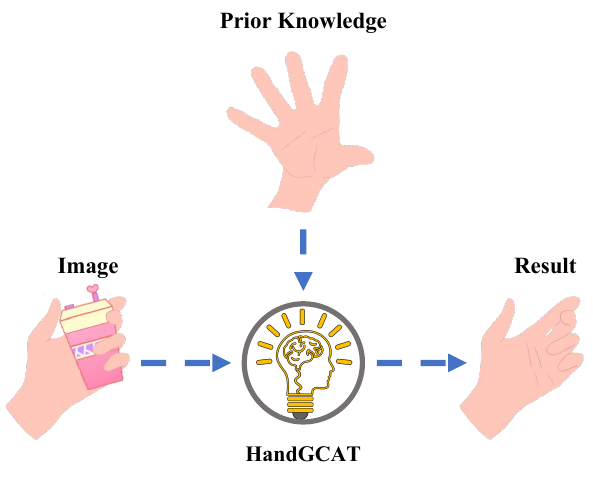}
  \caption{
  The main idea of the proposed HandGCAT is to exploit the hand prior knowledge to imagine occluded regions.
  }
  \label{fig1}
\end{figure}
\begin{figure*}
  \centering
  \includegraphics[width=\linewidth]{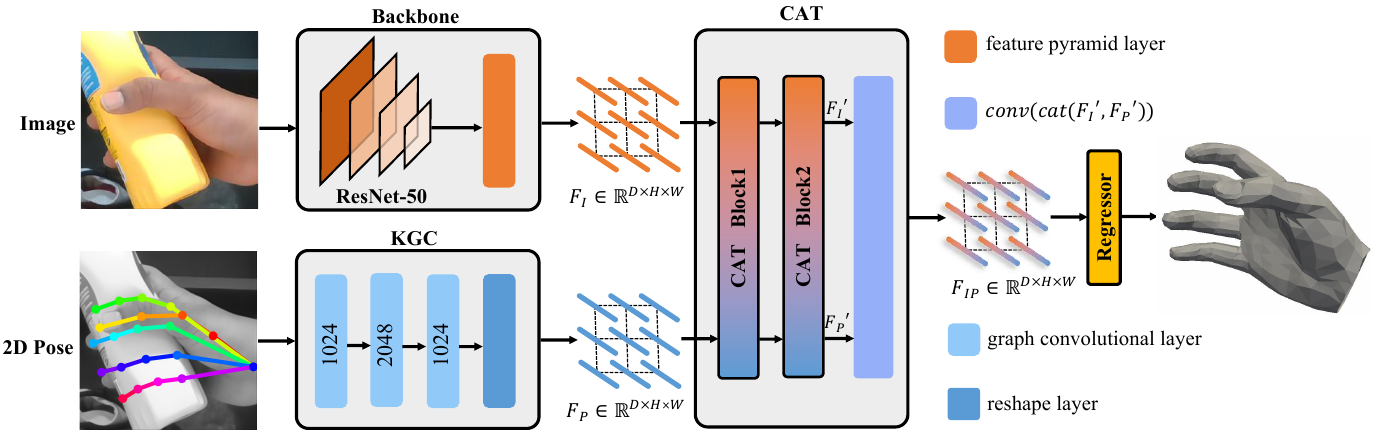}
  \caption{
  Overview of the proposed HandGCAT for 3D hand mesh reconstruction, which consists of backbone, KGC, CAT, and regressor. Resnet-50 with FPN extracts image feature $F_I$. KGC captures hand prior knowledge $F_P$ using GCNs from the 2D pose. CAT fuses $F_P$ into $F_I$ and thus imagines occluded regions. Finally, the regressor reconstructs the 3D hand mesh.
  }
\label{fig2}
\end{figure*}

Previous methods \cite{sarandi2018robust, data_aug_1,data_aug_2} have attempted to alleviate occlusion problem through data augmentation.
However, these augmentation methods are limited in the wild due to significant differences between synthetic and real occlusions.
More recently, \cite{occlusion_aware, handoccnet} exploited the attention mechanism to remove the interference of occlusions.
Although the attention-based methods have shown noticeable results under the occlusions, there are several limitations. First, these methods cannot imagine occluded regions. Second, they struggle to distinguish cluttered backgrounds that are unrelated to hands.

To resolve the above issues, let's think carefully about how humans solved this.
Humans can easily imagine the appearance of the hand in occluded regions since they know exactly what the hand looks like.
Inspired by this, we propose the HandGCAT, a novel framework for occlusion-robust 3D hand mesh reconstruction. 
As shown in Fig. \ref{fig1}, the main idea of the HandGCAT is to imagine occluded regions with prior knowledge.
The main components of our network include a Knowledge-Guided Graph Convolution (KGC) module and a Cross-Attention Transformer (CAT) module.
Guided by a predefined hand skeleton, KGC learns to hand prior knowledge from 2D hand pose based on a Graph Convolutional Network (GCN).
CAT leverages hand prior knowledge to find relevant occluded regions and fuses prior information into occluded regions, which allows the network to imagine occluded regions as humans do.
It has two advantages over existing methods.
First, HandGCAT exploits 2D hand prior knowledge to compensate for the missing information in occluded regions. In addition, 2D hand pose can be estimated accurately from color images since many well-performing methods \cite{alphapose,openpose} are trained on real large-scale datasets.
Second, HandGCAT fuses image features and hand prior knowledge for reconstruction, which can effectively alleviate the overfitting related to image appearance.
For evaluation, we report the model's performance on three challenging benchmarks containing severe occlusions: HO3D v2 \cite{HO3D_v2}, HO3D v3 \cite{HO3D_v3}, and DexYCB \cite{dex_ycb}.
Without whistles and bells, the HandGCAT can outperform the results of state-of-the-art methods.

Our main contributions are summarized as follows: (1) We propose a novel framework, HandGCAT, that recovers 3D hand mesh from a single RGB image. It is robust to severe hand occlusions and free from overfitting to image appearance.
(2) To obtain hand prior knowledge and fuse it into occluded regions, We propose KGC and CAT modules. Based on a predefined hand skeleton, the KGC leverages the GCN to model the interactions of 2D hand joints. The CAT adopts hand prior to compensate for hand occluded regions and improves the accuracy of reconstructing hand mesh under occlusions.
(3) Extensive experiments show our framework significantly outperforms state-of-the-art 3D hand mesh reconstruction methods on hand-object interaction datasets that contain severe hand occlusions.

\section{Related works}

\subsection{3D Hand Mesh Reconstruction}
Most of the 3D hand mesh reconstruction methods use a parametric model such as MANO \cite{mano} to represent the hand pose and shape.
\cite{data_aug_1,data_aug_2} utilized data augmentation during the training time to handle the occlusion.
\cite{occlusion_aware, handoccnet} applied the attention mechanism to improve the robustness under occlusions.
In this work, we propose a knowledge-guided occlusion robustness strategy that achieves state-of-the-art reconstruction results compared to previous methods.

\subsection{Graph Convolutional Networks}
Compared with traditional Convolutional Neural Networks (CNN), Graph Convolutional Networks (GCN) can perform convolutional operations on graph data.
As the skeleton can be represented in the form of a graph, GCNs are naturally widely used in 3D mesh reconstruction.
\cite{3d_gcn,pose2mesh_gcn} adopted GCN to model the topological relationships between mesh vertices and achieved state-of-the-art performance.
In this paper, we propose a knowledge-guided graph convolution module based on GCN to learn the implicit topology of the hand.

\subsection{Transformers}
Recently, many Transformer-based methods have been proposed for reconstructing 3D hand mesh from RGB images.
\cite{metro} proposed Transformer-based networks to model global vertex-to-vertex interactions.
\cite{mesh_graphormer} adopted a Transformer along with GCN to jointly model vertex-to-vertex and vertex-to-joint interactions.
However, how to effectively exploit the hand's implicit topology to compensate for occluded regions has rarely been explored.
In contrast, we propose a cross-attention Transformer that fuses hand features and implicit topology representations.

\section{HandGCAT}

\subsection{Overview}
The network architecture is shown in Fig. \ref{fig2}.
Given a hand image $I \in \mathbb{R}^{256\times 256 \times 3}$, Resnet-50 with FPN is used as a feature extractor to obtain a feature map $F_{I} \in \mathbb{R}^{256 \times 32\times 32}$.
Then, the KGC module takes the 2D pose $P_{2D} \in \mathbb{R}^{21\times 2}$ as input to perform joint-to-joint modeling, resulting in hand prior feature $F_{P} \in \mathbb{R}^{21\times 32\times 32}$ with rich topologic information.
Note that in training time, we adopt the ground truth 2D pose, and in testing time, the 2D pose is estimated from a pretrained keypoint detector \cite{handoccnet}.
However, the estimated 2D pose often contains errors. To enhance robustness to errors, we added randomly generated noise to the input ground truth 2D pose.
In the occlusion-aware stage, the CAT module fuses the corresponding hand prior into occluded regions, thus imagining the hand occluded regions, and finally outputs fused feature $F_{IP} \in \mathbb{R}^{256 \times 32\times 32}$.
In the reconstruction stage, the regressor takes the fused features $F_{IP}$ as input to predict the 3D vertex coordinates.

\begin{figure}
  \centering
  \includegraphics[width=0.8\linewidth]{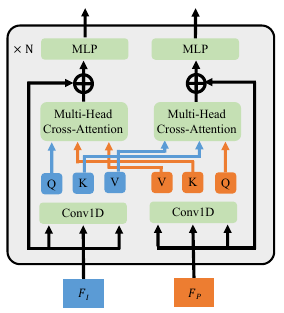}
  \caption{
  Cross-Attention Transformer Block.
  }
\label{fig3}
\end{figure}

\subsection{Knowledge-Guided Graph Convolution Module}
Humans have strong knowledge extraction and commonsense reasoning abilities. Therefore humans can easily imagine hand occluded regions.
Inspired by this, we designed a Knowledge-Guided Graph Convolution module, which learns hand prior from 2D pose as complementary information for occluded regions. \\
\textbf{Graph construction.} 
We construct an undirected graph $\mathcal{G} = (\mathcal{V}, W)$ of $P_{2D}$, where $\mathcal{V}=\{p_{i}\}_{i=1}^{J}$ denotes a set of $J=21$ hand joints, $W =(w_{i,j})_{J \times J}$ is the adjacency matrix. 
$W$ defines the edge connections between joints based on the hand skeleton \cite{mano}, where $w_{i,j}=1$ if joint $i$ and joint $j$ are connected, and $w_{i,j}=0$ otherwise.
The normalized graph Laplacian is computed as $L_P=I_J-D^{-\frac{1}{2}} W D^{-\frac{1}{2}}$, where $I_J$ is the identity matrix, $D$ is the degree matrix, and $D_{i i}=\sum_j W_{i j}$.
The scaled Laplacian \cite{pose2mesh_gcn} is computed as $\tilde{L_{P}}=2L_P/\lambda_{max}-I_J$.
\\
\textbf{Graph convolution on pose.}
The KGC module performs graph convolution operations on the 2D pose $P_{2D}$ based on Chebysev spectral graph convolution \cite{chebysev}, modeling joint and joint interactions to obtain hand prior information $F_{P}$.
The formulation for performing the graph convolution is as follows.
\begin{equation}
F_{\text {out }}=\sum_{k=0}^{K-1} T_k\left(\tilde{L_{\mathrm{P}}}\right) F_{\mathrm{in}} \Theta_k    
\end{equation}
where $T_{k}$ is the Chebyshev polynomial of order $k$.
We set K=1 in the KGC module, which means 1-hop neighbor nodes from each node are affected \cite{pose2mesh_gcn}.
$F_{in} \in \mathbb{R}^{J\times f_{in}}$ and $F_{out} \in \mathbb{R}^{J\times f_{out}}$ are the input and output feature maps with dimensions $f_{in}$ and $f_{out}$, respectively.
$\Theta_k \in \mathbb{R}^{f_{in}\times f_{out}}$ is a vector of Chebyshev coefficients.
\\

\begin{table*} \footnotesize \centering \renewcommand\arraystretch{1.25}
\caption{Comparison with state-of-the-art methods on HO3D v2.}
\begin{tabular}{ccccccccc}
\toprule
\tabcolsep=1cm
Method & PA-MPJPE $\downarrow$ & PA-MPJPE AUC $\uparrow$ & PA-MPVPE $\downarrow$ & PA-MPVPE AUC $\uparrow$ & F@5 $\uparrow$ & F@15 $\uparrow$ \\
\midrule
I2L-MeshNet \cite{i2l_meshnet} (CVPR’20)&  11.2  &  0.775  &  13.9  &  0.722  &  0.409  &  0.932 \\
Hasson \textit{et al}. \cite{hasson2020leveraging} (CVPR’20)& 11.0  &  0.780  &  11.2  &  0.777  &  0.464  &  0.939 \\
Hampali \textit{et al}. \cite{hampali2020honnotate} (CVPR’20)&  10.7  &  0.788  &  10.6  &  0.790  &  0.506  &  0.942 \\
METRO \cite{metro} (CVPR’21)&  10.4  &  0.792  &  11.1  &  0.779  &  0.484  &  0.946 \\
Liu \textit{et al}. \cite{liu2021semi} (CVPR’21)&  9.9  &  0.803  &  9.5  &  0.810  &  0.528  &  0.956 \\
I2UV-HandNet \cite{i2uv_handnet} (ICCV’21)&  9.9  & 0.804  &  10.1  &  0.799  &  0.500  &  0.943 \\
ArtiBoost \cite{artiboost} (CVPR’22)& 11.4  &  0.773  &  10.9  &  0.782  &  0.488  &  0.944 \\
Keypoint Trans. \cite{keypoint_transformer} (CVPR’22)&  10.8  &  0.786  &  -  &  -  &  -  &  -  \\
MobRecon \cite{MobRecon} (CVPR’22)&  9.2  &  -  &  9.4  &  -  &  0.538  &  0.957 \\
HandOccNet \cite{handoccnet} (CVPR’22)&  9.1  &  0.819  &  8.8  &  0.819  &  0.564  &  \textbf{0.963} \\
\midrule
\textbf{HandGCAT (Ours)} &  \textbf{8.7}  &  \textbf{0.826}  &  \textbf{8.7}  &  \textbf{0.827}  &  \textbf{0.584}  &  \textbf{0.963} \\
\bottomrule
\end{tabular}
\label{tab1}
\end{table*}

\begin{table*} \footnotesize \centering \renewcommand\arraystretch{1.25}

\caption{Comparison with state-of-the-art methods on HO3D v3.}
\begin{tabular}{ccccccccc}
\toprule
\tabcolsep=1cm
Method & PA-MPJPE $\downarrow$ & PA-MPJPE AUC $\uparrow$ & PA-MPVPE $\downarrow$ & PA-MPVPE AUC $\uparrow$ & F@5 $\uparrow$ & F@15 $\uparrow$ \\
\midrule
ArtiBoost \cite{artiboost} (CVPR’22) & 10.8 & 0.785 & 10.4 & 0.792 & 0.507 & 0.946\\
Keypoint Trans. \cite{keypoint_transformer} (CVPR’22)& 10.9 & 0.785 & - & - & - & -\\
HandOccNet \cite{handoccnet} (CVPR’22)& 10.7 & 0.786 & 10.4 & 0.791 & 0.479 & 0.935\\
\midrule
\textbf{HandGCAT (Ours)} & \textbf{9.3} & \textbf{0.814} & \textbf{9.1} & \textbf{0.818} & \textbf{0.552} & \textbf{0.956}\\
\bottomrule
\end{tabular}
\vspace{-2mm}
\label{tab2}
\end{table*}

\subsection{Cross-Attention Transformer Module}

In contrast to the conventional Transformer \cite{transformer}, we designed a Cross-Attention Transformer (CAT) module to imagine hand occluded regions with prior knowledge.
As shown in Fig. \ref{fig3}, CAT utilizes multiple stacked CAT blocks to implicitly model the correlation between hand prior $F_{P}$ and image feature $F_I$ with occlusions, potentially transferring $F_{P}$ to $F_{I}$.

We extract $Q_x, K_x, V_x, x \in \{I, P\}$ from $F_I$ and $F_P$ by two $1\times1$ convolution layers.
Following the position encoding (PE) in Transformer \cite{transformer}, position information is added to $Q_x$ and $K_x$:
\begin{equation}
    \begin{aligned}
    & Q_x = Q_x + PE(Q_x), \\
    & K_x = K_x + PE(K_x), 
    \end{aligned}
\end{equation}

Then, we use a multi-head cross-attention mechanism to perform the message passing between $F_I$ and $F_P$. 
The bidirectional interaction of Q and K, V between $F_I$ and $F_P$ allows CAT to efficiently mine strong correlations between $F_I$ and $F_P$.
This process is denoted as follows:

\begin{equation}
\begin{aligned}
&F_{I \rightarrow P}=Q_P + \operatorname{softmax}\left(\frac{Q_P K_I^T}{\sqrt{d}}\right) V_I, \\
&F_{P \rightarrow I}=Q_I + \operatorname{softmax}\left(\frac{Q_I K_P^T}{\sqrt{d}}\right) V_P,
\end{aligned}
\end{equation}
where $I\rightarrow P$ denotes pass $F_I$ to $F_P$, $P\rightarrow I$ otherwise, d is the scaling factor.
Afterward, we enhance the representation of the features using a multi-layer MLP as follows:
\begin{equation}
\begin{aligned}
&F_I^{\prime}=MLP\left(F_I+F_{P \rightarrow I}\right), \\
&F_P^{\prime}=MLP\left(F_P+F_{I \rightarrow P}\right),
\end{aligned}
\end{equation}

$F_I^{\prime}$ and $F_P^{\prime}$ can be taken as input for the next CAT block. After deep stacking multiple CAT blocks, we finally utilize a 1×1 convolutional layer to obtain the fused features $F_{IP}$.
\begin{equation}
F_{IP} = Conv1D(concat(F_I^{\prime}, F_P^{\prime}))
\end{equation}

\subsection{Regressor and Loss Function}
We follow the regressor from previous work \cite{liu2021semi}.
The regressor consists of a single hourglass network \cite{hourglass}, a parametric regression network, and a MANO layer.
The hourglass network takes the fused features $F_{IP}$ as input and outputs heatmaps $\mathbf{H} \in \mathbb{R}^{256 \times 32 \times 32}$ for each joint.

The parametric regression network concatenates the fused features $F_{IP}$ with heatmaps as input and outputs MANO pose $\theta \in \mathbb{R}^{48 \times 3}$ and MANO shape $\beta \in \mathbb{R}^{10}$ parameters.
The MANO layer produces 3D joints $J_{3D} \in \mathbb{R}^{21 \times 3}$ and 3D Mesh $V \in \mathbb{R}^{778 \times 3}$ based on MANO parameters.

To train our HandGCAT, we compute a loss function, which is defined as the $L_2$ distance between prediction $(\hat{\mathbf{H}},\hat{\theta},\hat{\beta},\hat{J_{3D}},\hat{V})$ and ground truth $(\mathbf{H},\theta,\beta,J_{3D},V)$.\\

\section{Experiments}

\subsection{Implementation Details}

All implementations were done with PyTorch.
Adam optimizer is adopted to train the network with a mini-batch size of 32. All models in our experiments are trained for 70 epochs. The initial learning rate is $10^{-4}$, which is scaled by a factor of 0.7 at every 10th epoch. All other details will be available in our codes.

\subsection{Datasets and Evaluation Metrics}

\textbf{HO3D v2}. 
HO3D v2 \cite{HO3D_v2} is a hand-object interaction dataset with severe occlusions, which contains 66,034 training samples and 11,524 testing samples. 
For the training set, each sample provides a single-view RGB image, MANO-based hand joints and mesh, and camera parameters. 
For the testing set, each sample has an RGB image with the annotation of the detected bounding box. 
The results of the testing set can be evaluated through an online submission system.
\\
\textbf{HO3D v3}.
Compared to HO3D v2, HO3D v3 \cite{HO3D_v3} has a larger scale, which provides 83,325 training samples and 20,137 testing samples. HO3D v3 provides more accurate annotations of hand-object poses, resulting in more contact between the hand and the object.
\\
\textbf{DexYCB}.
DexYCB \cite{dex_ycb} is a large-scale dataset of object grasping, containing 582K image frames of grasping on 20 YCB objects.
We make use of the official "S0" split to evaluate the right-hand pose.
\\
We report evaluation results using the following metrics.
\textbf{MPJPE/MPVPE} measures the average Euclidean distance (in mm) between the predicted joint/vertex coordinates and the ground truth coordinates.\\
\textbf{PA-MPJPE/MPVPE} is obtained by modifying MPJPE/MP- VPE using Procrustes analysis.\\
\textbf{AUC} means the Area Under Curve (AUC) of the Percentage of Correct Keypoints (PCK) and Vertices (PCV) at different error thresholds.
\\
\textbf{F-Score} is the harmonic mean between the predicted mesh and ground truth mesh at a specific distance threshold follow \cite{liu2021semi}. F@5/F@15 corresponds to the threshold of 5mm/15mm, respectively.
\vspace{1mm}
\subsection{ Comparisons with the State-of-the-art Methods}
We compared the proposed method with several state-of-the-art methods to reconstruct hand mesh on the HO3D v2 and HO3D v3 datasets.
In Tab. \ref{tab1}\&\ref{tab2}, we can see that the proposed method outperforms the previous state-of-the-art for both datasets.
To further validate the robustness of our method to occlusions, we compare it with state-of-the-art methods on a larger dataset DexYCB in Tab. \ref{tab3}. 
Our method also outperforms all previous methods. 
In summary, the experimental results on three datasets all demonstrate the generalization capability of the proposed method to the situation where the hand is severely occluded.
\vspace{1mm}
\subsection{Ablation Study}
To prove the effectiveness of the HandGCAT method, we conduct further analysis on several essential components.
The experimental results on the KGC module and CAT module are reported in Tab. \ref{tab4}\&\ref{tab5}, respectively.
\begin{table}[H] \footnotesize \centering 
\renewcommand\arraystretch{1.25}
\caption{Comparison with SOTA on DexYCB dataset.}
\begin{tabular}{ccccccccc}
\toprule
\tabcolsep=1cm
Method & MPJPE $\downarrow$ & PA-MPJPE $\downarrow$ \\
\midrule
Spurr \textit{et al}. \cite{spurr2020weakly} (ECCV’20) & 17.34 & 6.83\\
METRO \cite{metro} (CVPR’21) & 15.24 & 6.99\\
Liu \textit{et al}. \cite{liu2021semi} (CVPR’21) & 15.28 & 6.58\\
HandOccNet \cite{handoccnet} (CVPR’22) & 14.04 & 5.80\\
\midrule
\textbf{HandGCAT (Ours)} & \textbf{13.76} & \textbf{5.60} \\
\bottomrule
\end{tabular}
\label{tab3}
\end{table}
\begin{table}[H] \footnotesize \centering \renewcommand\arraystretch{1.25}
\vspace{-2mm}
\caption{Comparison of models with various KGC architectures on
HO3D v2.}
\resizebox{\linewidth}{!}{
\begin{tabular}{ccccccccc}
\toprule
\tabcolsep=1cm
KGC architectures & PA-MPJPE $\downarrow$ & PA-MPVPE $\downarrow$ & F@5 $\uparrow$& F@15 $\downarrow$ \\
\midrule
MLP & 9.3 & 9.3 & 0.547 & 0.959 \\
1-layer GCN & 9.2 & 9.2 & 0.546 & 0.961 \\
2-layer GCNs & 9.0 & 8.9 & 0.570 & 0.961 \\
3-layer GCNs & 8.9 & 8.8 & 0.573 & \textbf{0.963} \\
4-layer GCNs & \textbf{8.7} & \textbf{8.7} & \textbf{0.584} & \textbf{0.963} \\
5-layer GCNs & 8.9 & 8.8 & 0.579 & 0.962 \\
\bottomrule
\end{tabular}}
\label{tab4}
\end{table}
\begin{table}[H] \footnotesize \centering \renewcommand\arraystretch{1.25}
\vspace{-2mm}
\caption{Comparison of models with various CAT architectures on
HO3D v2.}
\resizebox{\linewidth}{!}{
\begin{tabular}{ccccccccc}
\toprule
\tabcolsep=1cm
CAT architectures & PA-MPJPE $\downarrow$ & PA-MPVPE $\downarrow$ & F@5 $\uparrow$ & F@15 $\uparrow$\\
\midrule
Two Transformers & 9.0 & 9.0 & 0.563 & 0.962 \\
Single CAT block & 8.9 & 8.8 & 0.574 & 0.962 \\
Two CAT blocks & \textbf{8.7} & \textbf{8.7} & \textbf{0.584} & \textbf{0.963} \\
Three CAT blocks & 8.8 & \textbf{8.7} & 0.583 & \textbf{0.963} \\
\bottomrule
\end{tabular}}
\label{tab5}
\end{table}
\begin{figure}
  \centering
  \includegraphics[width=0.8\linewidth]{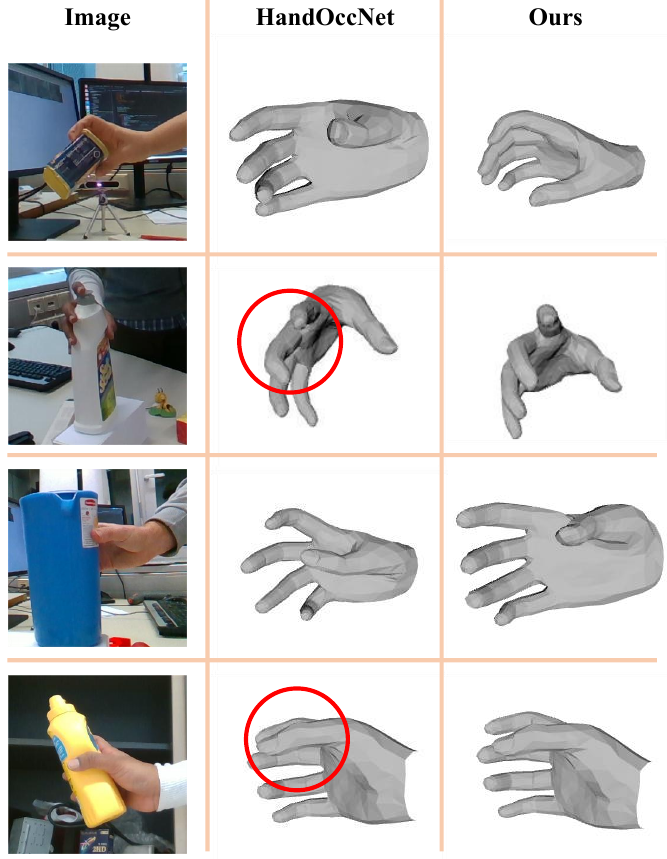}
  \caption{
  Qualitative comparison of the proposed HandGCAT and state-of-the-art method \cite{handoccnet} on HO3D v2.
  }
\label{fig4}
\end{figure}

\textbf{Effectiveness of the KGC module.}
Tab. \ref{tab4} shows that our KGC module with 4-layer GCNs achieves the best results in all metrics, where MLP represents a 3-layer perceptron. 
By observing the Tab. \ref{tab4}, we can see that: (1) When using MLP instead of the KGC module, the error of the reconstruction is dramatically raised. This is because MLP struggles in learning the hand skeleton spatial structure. It demonstrated that the KGC module performs essential roles in learning 2D prior knowledge.
(2) The KGC module achieves the best results with 4-layer GCN, and degradation occurs with 5-layer GCN.
This is due to the KGC module with shallow GCN layers (\textit{e.g.}, 1, 2, and 3 layers) being unable to model long-range joint interactions. However, the over-smooth problem of deeper GCN layers (\textit{e.g.}, 5 layers) leads to performance degradation.

\textbf{Effectiveness of the CAT module.}
Tab. \ref{tab5} shows that our CAT module with two stacked CAT blocks achieves the best results in all metrics, where two transformers represent two layers of the traditional transformer. 
From Tab. \ref{tab5}, we list the following observations: (1) With the transformer structure instead of the CAT module, which achieves slightly worse results. This suggests that the traditional transformer fails to capture the high correlation between prior knowledge and image features.
(2) Two stacked CAT blocks are more effective than a single CAT block and three CAT blocks. This is because a single CAT block cannot adequately enhance the image feature, and an additional CAT block has a marginal effect on compensating for information in occluded regions.
\\

\subsection{Qualitative Analysis}
Qualitative results on HO3D v2 are shown in Fig. \ref{fig4}. 
It can be seen that our HandGCAT produces the 3D mesh much better than the state-of-the-art method HandOccNet \cite{handoccnet}, even under severe occlusions.
In the first example, the results generated by HandOccNet \cite{handoccnet} are completely off from the angle of the original RGB image. In contrast, HandGCAT predicted relatively accurate results. 
With the help of hand prior knowledge, HandGCAT successfully imagined occluded regions, thus improving the performance of reconstruction.

\section{Conclusion}
In order to mitigate the challenges posed by hand occlusions, we propose a novel hand reconstruction method named HandGCAT, which is robust to occlusions.
First, we propose a knowledge-guided graph convolution module to learn the hand prior.
Second, we design a cross-attention Transformer to fuse the hand prior into occluded regions.
Extensive experimental results show that our method achieves state-of-the-art performance on 3D hand mesh benchmarks that contain severe occlusions.

\section{Acknowledgment}
This work is supported by National Key R\&D Program of China (2021ZD0113502), Shanghai Municipal Science and Technology Major Project (2021SHZDZX0103).
\clearpage

\bibliographystyle{IEEEbib}

\end{document}